%% file: sigdial17_Diligent_short.tex
\documentclass[11pt,a4paper]{article}
\usepackage[nohyperref]{acl2017} 
\usepackage{times}
\usepackage{latexsym}
\usepackage{adjustbox}
\usepackage{url}
\usepackage{graphicx}
\usepackage{adjustbox}
\usepackage{caption}
\usepackage{subcaption}
\usepackage{multirow}

\newcommand{\ignore}[1]{}

\usepackage{amsmath}
\usepackage{xcolor}
\usepackage[normalem]{ulem}

\def\JNdel#1{\bgroup\markoverwith{\textcolor{blue}{\rule[0.5ex]{2pt}{1pt}}}\ULon{#1}}

\definecolor{darkgreen}{rgb}{0.0, 0.5, 0.0}

\makeatletter
\def\ODhl#1{\bgroup \markoverwith{\lower3.5\p@\hbox{\sixly \textcolor{darkgreen}{\char58}}}\ULon{#1}}
\font\sixly=lasyb10 scaled 652 
\makeatother
\def\ODdel#1{\bgroup\markoverwith{\textcolor{darkgreen}{\rule[0.5ex]{2pt}{1pt}}}\ULon{#1}}

\aclfinalcopy 
\def\aclpaperid{71} 

\title{The E2E NLG Shared Task}
\title{The E2E Dataset: New Challenges For End-to-End Generation}

\author{Jekaterina Novikova, Ond\v{r}ej Du{\v{s}}ek and Verena Rieser \\
  School of Mathematical and Computer Sciences \\
  Heriot-Watt University, Edinburgh \\
  {\tt j.novikova, o.dusek, v.t.rieser@hw.ac.uk} 
  }
  
\date{}

\begin{document}
\maketitle
\begin{abstract}
This paper describes the E2E data, a new dataset for training end-to-end, data-driven natural language generation systems in the restaurant domain, which is ten times bigger than existing, frequently used datasets in this area. The E2E dataset poses new challenges: (1) its human reference texts show more lexical richness and syntactic variation, including discourse phenomena; (2) generating from this set requires content selection. As such, learning from this dataset promises more natural, varied and less template-like system utterances.
We also establish a baseline on this dataset, which illustrates some of the difficulties associated with this data.
\end{abstract}

\section{Introduction}
The natural language generation (NLG) component of a spoken
dialogue system typically has to be re-developed for every new application domain.
Recent end-to-end, data-driven NLG systems, however, promise rapid development of NLG components in new domains:
They jointly learn sentence planning and surface realisation from non-aligned data (\citealp{jurcicek:2015:ACL,wen:emnlp2015,Mei:NAACL2016,Wen:NAACL16}; \citeauthor{SharmaHSSB16}, \citeyear{SharmaHSSB16}; \citealp{Dusek:ACL16,vlachos:coling2016}).
These approaches do not require costly semantic alignment between meaning representations (MRs) and the corresponding natural language (NL) reference texts (also referred to as ``ground truths" or ``targets"), but they are trained on parallel datasets, which can be collected in sufficient quality and quantity using effective crowdsourcing techniques, e.g.\ \cite{novikova:INLG2016}. 
%
So far, end-to-end approaches to NLG are limited to small, delexicalised datasets, e.g.\ BAGEL \cite{mairesse:acl2010}, SF Hotels/\hspace{0mm}Restaurants \cite{wen:emnlp2015}, or RoboCup \cite{chen2008learning}.
Therefore, end-to-end methods have not been able to replicate the rich dialogue and discourse phenomena targeted by previous rule-based and statistical  approaches  for language generation in dialogue, e.g.\ \cite{walker2004generation,stent2004trainable,demberg2006information,rieser2009natural}. 

In this paper, we describe a new crowdsourced dataset of 50k instances in the restaurant domain (see Section~\ref{sec:data}). We analyse it following the methodology 
proposed by \citet{Perez-Beltrachini17} and show that the dataset brings additional  
challenges, such as open vocabulary, complex syntactic structures and diverse discourse phenomena, as described in Section~\ref{sec:challenges}.
The data is openly released as part of the E2E NLG challenge.\footnote{\url{http://www.macs.hw.ac.uk/InteractionLab/E2E/}}
We establish a baseline on the dataset  in Section~\ref{sec:system}, using one of the previous end-to-end approaches. 

\section{The E2E Dataset}\label{sec:data}

\begin{table}[tb]
\footnotesize
\centering
\begin{adjustbox}{max width=0.48\textwidth}
\begin{tabular}{ll}
\bf Flat MR & \bf NL reference \\
\hline \hline
\multirow{3}{*}
{\begin{tabular}[c]{@{}l@{}}name{[}Loch Fyne{]}, \\ eatType{[}restaurant{]}, \\ food{[}French{]}, \\ priceRange{[}less than \pounds20{]}, \\ familyFriendly{[}yes{]}\end{tabular}} & \begin{tabular}[c]{@{}l@{}}Loch Fyne is a family-friendly\\ restaurant providing wine and\\ cheese at a low cost.\end{tabular} \\ \\
 & \begin{tabular}[c]{@{}l@{}}Loch Fyne is a French family\\ friendly restaurant catering to\\ a budget of below \pounds20.\end{tabular} \\ \\
 & \begin{tabular}[c]{@{}l@{}}Loch Fyne is a French\\ restaurant with a family setting\\ and perfect on the wallet.\end{tabular}
\end{tabular}
\end{adjustbox}
\caption{An example of a data instance. 
}
\label{tab:1}
\end{table}
\normalsize

 \begin{figure}[ht]
 \centering
  \includegraphics[width=.9\linewidth]{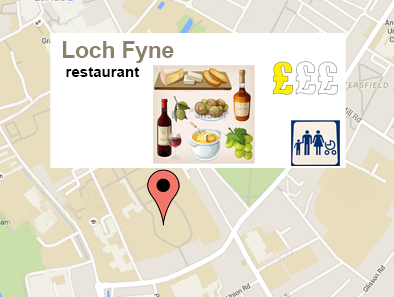}
 \caption{Pictorial MR for Table \ref{tab:1}. } 
 \label{fig:picture}
 \end{figure}


 \begin{table}[tb]
\begin{center}
\begin{adjustbox}{max width=0.48\textwidth}
\begin{tabular}{lll}
\textbf{Attribute} & \textbf{Data Type} & \textbf{Example value}\\ \hline \hline
name & verbatim string & The Eagle, ...\\  
eatType & dictionary & restaurant, pub, ...\\  
familyFriendly & boolean & Yes / No\\  
priceRange & dictionary & cheap, expensive, ...\\  
food & dictionary & French, Italian, ...\\  
near & verbatim string & market square, ... \\ 
area & dictionary & riverside, city center, ...\\  
customerRating & enumerable & 1 of 5 (low), 4 of 5 (high), ...\\  
\end{tabular}
\end{adjustbox}
\end{center}
\caption{Domain ontology of the E2E dataset.}
\label{tab:attr}
\end{table}%

\begin{table*}[tb]
\begin{center}
\begin{tabular}{lrrrcrrrr}
        & \begin{tabular}[c]{@{}c@{}}No. of \\ instances\end{tabular} & \begin{tabular}[c]{@{}c@{}}No. of \\ unique MRs\end{tabular} & \multicolumn{2}{c}{Refs/MR} & \hspace{-3mm}Slots/MR & W/Ref & W/Sent   & Sents/Ref \\
\hline \hline
E2E  & \textbf{50,602}            & \textbf{5,751}          & \textbf{8.1\phantom{0}}\hspace{-2mm} & \textbf{(2--16)\phantom{0}}   & \textbf{5.43} & \textbf{20.1\phantom{0}}      & \textbf{14.3\phantom{0}}  & \textbf{1.5\phantom{0} (1--6)}   \\
SFRest  & 5,192             & 1,950          & 1.82\hspace{-2mm} & (1--101) & 2.86 & 8.53       & 8.53  & 1.05 (1--4)         \\
Bagel   & 404              & 202           & 2\phantom{.00}\hspace{-2mm} & (2--2)\phantom{00} & 5.41 & 11.54      & 11.54 & 1.02 (1--2)         
\end{tabular}
\end{center}
\caption{Descriptive statistics of linguistic and computational adequacy of datasets.}

\medskip\small
\textit{No. of instances} is the total number of instances in the dataset, \textit{No. of unique MRs} is the number of distinct MRs, \textit{Refs/MR} is the number of NL references per one MR (average and extremes shown), \textit{Slots/MR} is the average number of slot-value pairs per MR, \textit{W/Ref} is the average number of words per MR, \textit{W/Sent} is the average number of words per single sentence, \textit{Sents/Ref} is the number of NL sentences per MR (average and extremes shown).
\label{tab:res}
\end{table*}

The data was collected
using the CrowdFlower platform and quality-controlled following \newcite{novikova:INLG2016}. 
The dataset 
 provides information about restaurants and 
  consists of more than 50k combinations of a dialogue-act-based 
  MR 
   and 8.1 references on average, as shown in Table~\ref{tab:1}. 
The dataset is split into training, validation and testing sets (in a 76.5-8.5-15 ratio), keeping a similar distribution of MR and reference text lengths and ensuring that MRs in different sets are distinct.
Each MR consists of 3--8 attributes (slots), such as \emph{name}, \emph{food} or \emph{area}, and their values. A detailed ontology of all attributes and values is provided in Table~\ref{tab:attr}. 
%
%
%
Following \citet{novikova:INLG2016}, the E2E data was collected using pictures as stimuli (see example in Figure~\ref{fig:picture}), which was shown to elicit significantly more natural,
more informative, and better phrased human references than textual MRs.

\section{Challenges}\label{sec:challenges}

Following \citet{Perez-Beltrachini17}, we describe several different dimensions of our dataset and compare them to the BAGEL and SF Restaurants (SFRest) datasets, which use the same domain. 

\paragraph{Size:}
Table~\ref{tab:res} summarises the main descriptive statistics of all three datasets. 
The E2E dataset is significantly larger than the other sets in terms of instances,
 unique MRs, and average number of human references per MR (Refs/MR).%
\footnote{Note that the difference is even bigger in practice as the Refs/MR ratio for the SFRest dataset is skewed: 
for specific MRs, e.g. {\em goodbye}, SFRest  has 
 up to 101 references.} 
While having more data with a higher number of references per MR makes the E2E data more attractive for statistical approaches, 
it is also more challenging than previous sets as it uses a larger number of sentences in NL references (Sents/Ref; up to 6 in our dataset compared to typical 1--2 for other sets) and a larger number of slot-value pairs in MRs (Slots/MR).
 It also contains sentences of about double the word length 
 (W/Ref) and longer sentences in references (W/Sent).

\begin{figure*}[tb]
\centering
    \begin{subfigure}[b]{0.49\textwidth}
        \includegraphics[width=\textwidth]{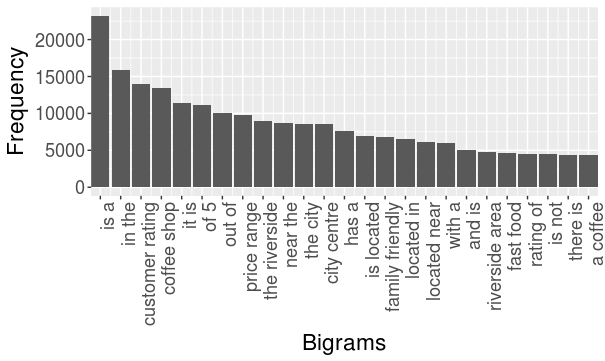}
    \end{subfigure}
    \hfill
    \begin{subfigure}[b]{0.49\textwidth}
        \includegraphics[width=\textwidth]{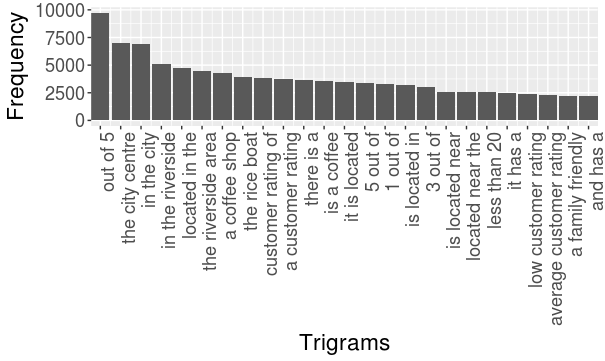}
    \end{subfigure}
\caption{Distribution of the top 25 most frequent bigrams and trigrams in our dataset (left: most frequent bigrams, right: most frequent trigrams).}
\label{fig:ngrams}
\vspace{-.5cm}
\end{figure*}

 \paragraph{Lexical Richness:}
 
\begin{table}[tb]
\centering
\begin{adjustbox}{max width=0.49\textwidth}
\begin{tabular}{lrrccc}
\textbf{Dataset} & \textbf{Tokens} & \textbf{Types} & \textbf{LS} & \textbf{TTR} & \textbf{MSTTR} \\
\hline\hline
E2E  & \textbf{65,710} & 945 & \textbf{0.57} & 0.01 & \textbf{0.75} \\
SFRest & 45,791 & 1,187 & 0.43 & 0.03 & 0.62 \\
Bagel & 1,071 & 70 & 0.42 & 0.04 & 0.41
\end{tabular}
\end{adjustbox}
\caption{Lexical Sophistication (LS) and Mean Segmental Type-Token Ratio (MSTTR).}
\label{tab:lexic}
\end{table}

We used the Lexical Complexity Analyser \cite{lu2012relationship} 
to measure various dimensions of lexical richness, 
as shown in Table~\ref{tab:lexic}. 
We complement the traditional measure of {\em lexical diversity}  {type-token ratio} (TTR) 
with the more robust measure of {mean segmental TTR} (MSTTR) \cite{lu2012relationship}, which divides the corpus into successive segments of a given length and then calculates the average TTR of all segments. The higher the value of MSTTR, the more diverse is the measured text.
Table~\ref{tab:lexic} shows our dataset has the highest MSTTR value (0.75) while Bagel has the lowest one (0.41).
In addition, we measure {\em lexical sophistication} (LS), also known as lexical rareness, 
which is calculated as the proportion of lexical word types not on the list of 2,000 most frequent words generated from the British National Corpus. 
Table~\ref{tab:lexic} shows that our dataset contains about 15\% more infrequent words 
compared to the other datasets.

We also investigate the distribution of the top 25 most frequent bigrams and trigrams in our dataset (see Figure~\ref{fig:ngrams}). The majority of both trigrams (61\%) and bigrams (50\%) is only used once in the dataset, which creates a challenge to efficiently train on this data. Bigrams used more than once in the dataset have an average frequency of 54.4 (SD = 433.1), and the average frequency of trigrams used more than once is 19.9 (SD = 136.9). For comparison, neither SFRest nor Bagel dataset contains bigrams or trigrams that are only used once. The minimal frequency of bigrams is 27 for Bagel (Mean  = 98.2, SD = 86.9) and 76 for SFrest (Mean  = 128.4, SD = 50.5), for trigrams the minimal frequency is 24 for Bagel (Mean = 63.5, SD = 54.6) and 43 for SFRest (Mean = 67.3, SD = 18.9). 
Infrequent words and phrases pose a challenge to current end-to-end generators since they cannot handle out-of-vocabulary words.


\begin{figure}[tb]
\centering
\includegraphics[width=.49\textwidth]{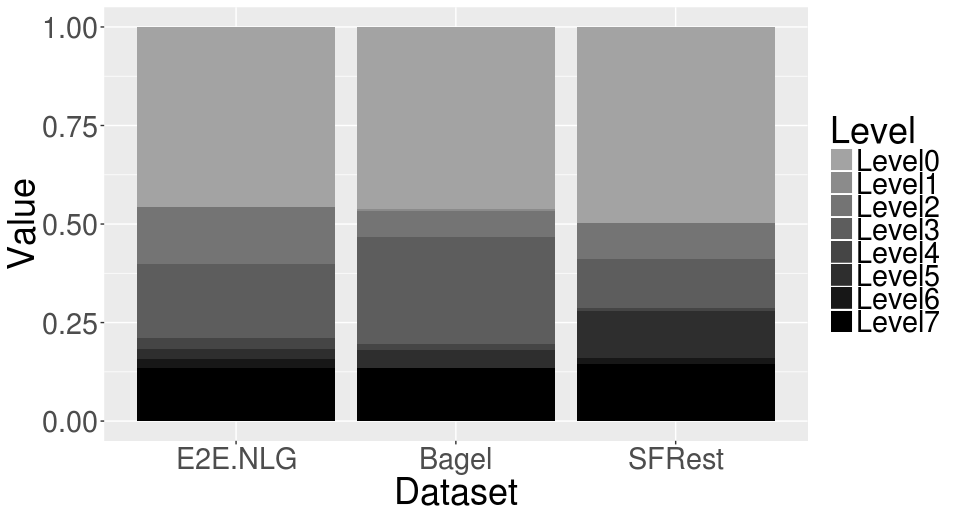}
\caption{
D-Level sentence distribution of the datasets under comparison. 
}
\label{fig:dlevels}
\end{figure}

 \paragraph{Syntactic Variation and Discourse Phenomena:}
We used the D-Level Analyser \cite{lu2009automatic} 
to evaluate syntactic variation and complexity of human references using the revised D-Level Scale \cite{lu2014computational}. 
Figure~\ref{fig:dlevels} show a similar syntactic variation in all three datasets. Most references in all the datasets are simple sentences (levels~0 and~1), although the proportion of simple texts is the lowest for the E2E NLG dataset (46\%) compared to others (47-51\%). Examples of simple sentences in our dataset include: ``The Vaults is an Indian restaurant", or ``The Loch Fyne is a moderate priced family restaurant". 
The majority of our data, however, contains more complex, varied syntactic structures, including phenomena explicitly modelled  by early statistical approaches \cite{stent2004trainable,walker2004generation}.
For example, clauses may be joined by a coordinating conjunction (level~2), e.g. ``Cocum is a very expensive restaurant \textit{but} the quality is great''. 
There are 14\% of level-2 sentences in our dataset, comparing to 7-9\% in others. 
Sentences may also contain verbal gerund (\emph{-ing}) phrases (level~4), either in addition to previously discussed structures or separately, e.g.\ ``The coffee shop Wildwood has fairly priced food, \textit{while being} in the same vicinity as the Ranch'' or ``The Vaults is a family-friendly restaurant \textit{offering} fast food at moderate prices''. 
Subordinate clauses are marked as level~5, e.g. ``\textit{If} you like Japanese food, try the Vaults''.
The highest levels of syntactic complexity involve sentences containing referring expressions (``The Golden Curry provides Chinese food in the high price range. \textit{It} is near the Bakers''), non-finite clauses in adjunct position (``\textit{Serving} cheap English food, as well as \textit{having} a coffee shop, the Golden Palace has an average customer rating and is located along the riverside'') or sentences with multiple structures from previous levels. All the datasets contain 13-16\% of sentences of levels~6 and~7, where Bagel has the lowest proportion (13\%) and our dataset the highest (16\%). 
 
\paragraph{Content Selection:}
In contrast to the other datasets, 
our crowd workers  
were asked to verbalise all the \textit{useful} information from the MR and were allowed to skip an attribute value  considered unimportant. This feature makes generating text from our dataset more challenging as NLG systems also need to learn which content to realise. 
In order to measure the extent of 
this phenomenon, we examined a random sample of 50 MR-reference pairs.
An MR-reference pair was considered a fully covered (C) match if all attribute values present in the MR are verbalised in the NL reference. It was marked
as ``additional" (A) if the reference contains information not present in the MR and as ``omitted" (O) if the MR contains information not present in the reference, see Table \ref{tab:semantics}.
40\% of our data contains either additional or omitted information. 
This often concerns the attribute-value pair \emph{eatType=restaurant}, which is either omitted (``Loch Fyne provides French food near The Rice Boat. It is located in riverside and has a low customer rating"
) or added in case \emph{eatType} is absent from the MR (``Loch Fyne is a low-rating riverside French \underline{restaurant} near The Rice Boat'').

\begin{table}[tb]
\begin{center}
\begin{tabular}{lccc}
\textbf{Dataset} & \textbf{O} & \textbf{A} & \textbf{C} \\
\hline \hline
E2E NLG & 22\% & 18\% & \phantom{0}60\% \\
SFRest & \phantom{0}0\% & \phantom{0}6\% & \phantom{0}94\% \\
Bagel & \phantom{0}0\% & \phantom{0}0\% & 100\% \\
\end{tabular}
\end{center}
\caption{Match between MRs and NL references.}  

\medskip\small
O: Omitted content, A: Additional content, C: Content fully covered in the reference.
\label{tab:semantics}
\end{table}





\section{Baseline System Performance}\label{sec:system}

To establish a baseline on the task data, we use TGen \cite{Dusek:ACL16}, one of the recent E2E data-driven systems.\footnote{TGen is freely available at \url{https://github.com/UFAL-DSG/tgen}.} TGen is based on  sequence-to-sequence modelling with attention (seq2seq) \cite{bahdanau_neural_2014}. 
In addition to the standard seq2seq model, TGen uses beam search for decoding and a 
reranker over the top $k$ outputs, penalizing those outputs that do not verbalize all attributes from the input MR.
As TGen does not handle unknown vocabulary well, 
the sparsely occurring string attributes (see Table~\ref{tab:attr}) \emph{name} and \emph{near} are delexicalized -- replaced with placeholders during generation time (both in input MRs and training sentences).\footnote{Detailed system training parameters are given in the supplementary material.}

\begin{table}[tb]
\begin{center}
\begin{tabular}{lc}
\bf Metric & \bf Value \\\hline\hline
BLEU \cite{papineni2002bleu} & 0.6925 \\ 
NIST \cite{nist}             & 8.4781 \\
METEOR \cite{meteor}         & 0.4703 \\
ROUGE-L \cite{lin2004rouge}  & 0.7257 \\
CIDEr \cite{cider}           & 2.3987 \\
\end{tabular}
\end{center}
\caption{TGen results on the development set.}\label{tab:metrics}
\end{table}

We evaluated TGen on the development part of the E2E set using several automatic metrics. The results are shown in Table~\ref{tab:metrics}.\footnote{To measure the scores, we used slightly adapted versions of the official MT-Eval script (BLEU, NIST) and the COCO Caption \cite{chen_microsoft_2015} metrics (METEOR, ROUGE-L, CIDEr). All evaluation scripts used here are available at \url{https://github.com/tuetschek/e2e-metrics}.}
Despite the greater variety of our dataset as shown in Section~\ref{sec:challenges}, the BLEU score achieved by TGen is in the same range as scores reached by the same system for BAGEL (0.6276) and SFRest (0.7270). This indicates that the size of our dataset and the increased number of human references per MR helps statistical approaches.

Based on cursory checks, generator outputs seem mostly fluent and relevant to the input MR. 
For example, our setup was able to generate long, multi-sentence output, including referring expressions and ellipsis, as illustrated by the following example:
``Browns Cambridge is a family-friendly coffee shop that serves French food. It has a low customer rating and is located in the riverside area near Crowne Plaza Hotel.''
However, TGen requires delexicalization and does not learn content selection, forcing the verbalization of all MR attributes.




\section{Conclusion}\label{sec:concl}

We described the E2E dataset for end-to-end, statistical natural language generation systems. While this dataset is ten times bigger than similar, frequently used datasets, it also poses new challenges given its lexical richness, syntactic complexity and discourse phenomena. Moreover, generating from this set also involves content selection. 
In contrast to previous datasets, the E2E data is crowdsourced using pictorial stimuli, which was shown to elicit more natural, more informative and better phrased human references than textual meaning representations \cite{novikova:INLG2016}. As such, learning from this data promises more natural and varied outputs than previous ``template-like" datasets.
The dataset is freely available as part of the E2E NLG Shared Task.
\footnote{The training and development parts of our dataset can be downloaded from \url{http://www.macs.hw.ac.uk/InteractionLab/E2E/}.}

In future work, we hope to collect data with further increased complexity, e.g.\ asking the user to compare, summarise, or recommend restaurants, in order to replicate previous rule-based and statistical approaches, e.g.\  \cite{walker2004generation,stent2004trainable,demberg2006information,Rieser:IEEE14}. In addition, we will experiment with collecting NLG data within a dialogue context, following \cite{duvsek-jurcicek:2016:SIGDIAL}, in order to model discourse phenomena across multiple turns.

\section*{Acknowledgements}
This research received funding from the EPSRC projects  DILiGENt (EP/M005429/1) and  MaDrIgAL (EP/N017536/1). The Titan Xp used for this research was donated by the NVIDIA Corporation.

\bibliography{acl2017}
\bibliographystyle{acl_natbib}

\input{supplementary}

\end{document}

%% file: supplementary.tex
\newpage
\onecolumn

\section*{The E2E Dataset Supplementary Material: Baseline Model Parameters}
\renewcommand\thetable{\Alph{table}} 

\begin{table}[h!]
\begin{center}
\begin{tabular}{lc}
\bf Setting & \bf Value \\\hline\hline
Adam optimizer learning rate & 5e-4 \\
Network cell type & LSTM \\
Embedding (+cell) size & 50 \\
Batch size & 20 \\
Encoder length (max.\ input attribute-value pairs) & 10 \\
Decoder length (max.\ output tokens) & 80 \\
Max. training epochs & 20 \\
Training instances reserved for validation & 2000 \\
\end{tabular}
\end{center}
\caption{TGen training parameters: main sequence-to-sequence (seq2seq) model with attention.}

\medskip
The training data are tokenized, lowercased, and values of \emph{name} and \emph{near} attributes are replaced with placeholders (``X-name'', ``X-near'') for training and generation.

The generator is trained by minimizing cross entropy in direct token-by-token generation of surface strings. Validation using BLEU on the reserved instances is performed after each epoch. Early stopping is in force: if top 3 BLEU results do not change for 5 epochs, training is finished. Network parameters giving best BLEU in validation are stored and used in the final model. We used 5 different random initializations of the model and selected the one that yielded best validation BLEU for final evaluation.
\end{table}

\begin{table}[h!]
\begin{center}
\begin{tabular}{lc}
\bf Setting & \bf Value \\\hline\hline
Adam optimizer learning rate & 1e-3 \\
Embedding (+cell) size & 50 \\
Batch size & 20 \\
Training epochs & 20 \\
Encoder length (max.\ input tokens) & 80 \\
Training instances reserved for validation & 2000 \\
\end{tabular}
\end{center}
\caption{TGen training parameters: reranker.}

\medskip
The reranker is trained to classify input MRs based on NL references (to be able to rerank main seq2seq generator outputs based on how well they reflect the input MR). Given that the \emph{name} and \emph{near} attributes are delexicalized, the result is a set of 21 classifiers (showing the presence or absence of each possible delexicalized attribute-value pair).

Validation on training and validation data is performed after each epoch and best-performing parameters are kept in the end. Validation data has 10 times more importance than training data for validation.
\end{table}

\begin{table}[h!]
\begin{center}
\begin{tabular}{lc}
\bf Setting & \bf Value \\\hline\hline
Beam size & 10 \\
Reranker misfit penalty & 100 \\
\end{tabular}
\end{center}
\caption{TGen decoding parameters.}

\medskip
The reranking penalty is very high so that outputs that best cover the input MR perfectly are always promoted to the top of the output 10-best list.  The correct values for the delexicalized attributes \emph{name} and \emph{near} are inserted in a simple postprocessing step.
\end{table}